\documentclass[letterpaper, 10 pt, conference]{ieeeconf}  

\IEEEoverridecommandlockouts                              

\usepackage[utf8]{inputenc}
\usepackage{graphics} 
\usepackage{epsfig} 
\usepackage{amsmath} 
\usepackage{hyperref}
\usepackage{subcaption}
\usepackage{booktabs}
\usepackage[font=small]{caption}
\usepackage{balance}

\usepackage{titlesec}
\let\labelindent\relax
\titlespacing{\subsection}{0pt}{*0.3}{*0.3}
\usepackage{enumitem}
\setlist[itemize]{noitemsep, ,nolistsep,topsep=0pt}
\setlist[enumerate]{noitemsep,nolistsep, topsep=0pt}
\title{\LARGE \bf
Audio-Visual Traffic Light State Detection for Urban Robots
}

\author{Sagar Gupta, Akansel Cosgun%
\thanks{Authors are with Deakin University, Australia}
}

\begin{document}

\maketitle
\thispagestyle{empty}
\pagestyle{empty}

\begin{abstract}


We present a multimodal traffic light state detection using vision and sound, from the viewpoint of a quadruped robot navigating in urban settings. This is a challenging problem because of the visual occlusions and noise from robot locomotion. Our method combines features from raw audio with the ratios of red and green pixels within bounding boxes, identified by established vision-based detectors. The fusion method aggregates features across multiple frames in a given timeframe, increasing robustness and adaptability. Results show that our approach effectively addresses the challenge of visual occlusion and surpasses the performance of single-modality solutions when the robot is in motion. This study serves as a proof of concept, highlighting the significant, yet often overlooked, potential of multi-modal perception in robotics.


\end{abstract}


\section{Introduction}

The development of urban mobile robots is bringing big changes, especially in areas like package delivery, surveillance, and as robotic helpers. For these robots to move around city streets safely and effectively, they need to be good at navigating autonomously, especially when moving on sidewalks or crossing streets. A key part of this navigation is being able to spot pedestrian traffic lights (PTLs) and knowing when the lights are red or green, to ensure safe and effective robot navigation in urban environments.

Recent advancements in deep learning have notably enhanced computer vision, leading to the creation of sophisticated vision-based object detectors \cite{yolo, he2017mask, tian2019fcos}. However, these systems often struggle with occlusions, a significant challenge in urban environments where pedestrians, bicycles, or vehicles can block the robot's camera view. This issue is more pronounced for ground robots with lower camera positions, intensifying the problem of occlusion. To overcome this, exploring other sensor modalities becomes crucial. Sound, for instance, can complement visual data effectively. The integration of audio and visual information has shown promising results in various fields, including deepfake detection \cite{muppalla2023integrating}, robot learning \cite{Du2022}, emotion recognition \cite{Jia2022AME} and multimedia analysis \cite{atrey2010multimodal}. Nonetheless, the application of audio-visual fusion for traffic light state detection remains largely unexplored.

\begin{figure}[ht]
\centering
\includegraphics[width=0.77\linewidth,trim={0 2cm 0 0},clip]{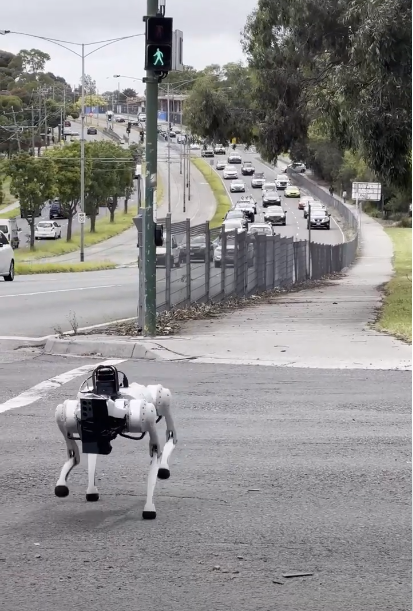}
\caption{In urban settings, robots often struggle with visual occlusion, hindering their ability to detect traffic lights. Our solution combines auditory cues with vision, utilizing traffic lights' sound patterns to indicate their state. This method ensures robots can navigate effectively, even when visual signals are obscured.}
\label{fig:intro}
\end{figure}

We address this research gap by focusing on PTL detection from the perspective of a quadruped robot navigating real urban environments. Our approach is particularly suited for PTLs that emit sound patterns corresponding to their light state, designed for aiding visually impaired pedestrians. Our vision-based detection (Section ~\ref{sec:vision}) utilizes a third-party object detector to get the traffic light housing bounding box, which then we use a simple pixel counting approach in the HSV color scale. Our audio-based detection (Section ~\ref{sec:audio}) takes raw audio as input, extracts Mel-Frequency Cepstral Coefficients (MFCC) as features \cite{mfccreview} and uses a random forest classifier \cite{biau2016random}. We propose a feature-level audio-visual fusion model (Section ~\ref{sec:fusion}) that incorporates a sequence of frames in a given timeframe. In Section \ref{sec:results}, we report the accuracy of our audio-visual PTL state classifier under the following conditions: 

\begin{itemize}
    \item When the robot is not moving, and under no occlusions
    \item When the robot is not moving, but its view is blocked
    \item When the robot is moving, and under no occlusions
\end{itemize}

Section \ref{sec:demo} details the implementation of our approach on a Unitree Go1 quadruped robot, showcasing it autonomously crossing the road when the light turns green\footnote{\href{https://www.youtube.com/watch?v=Wm293-VgKzI}{https://www.youtube.com/watch?v=Wm293-VgKzI}}. The contribution of this paper is three-fold:

\begin{itemize}
\item An audio-visual fusion model that uses audio features and color histograms in given bounding boxes in a given sequence of frames. 
\item The first time audio-visual fusion is applied to traffic light state detection for urban robots.
\item Analysis on a dataset taken by a quadruped robot, under varying visual occlusion and robot motion.
\end{itemize}




\section{System Overview}

The system, designed for urban robots, classifies PTLs as red or green by integrating both visual and auditory data. It comprises three primary components: vision-based detection, audio-based detection, and audio-visual fusion. For model training, we used two datasets. The visual dataset comprised images from the ImVisible dataset \cite{yu2019lytnet}, featuring PTLs from Shanghai intersections. This dataset was enriched with manual annotations of normalized coordinates for PTL bounding boxes in YOLO format, into two classes: red or green. Our auditory dataset included two hours of handheld audio recordings each for red and green class. These recordings were split into two audio files, each containing data for a distinct class, either red or green. The audio data was collected using a smartphone microphone from 3 crosswalks in Melbourne, Australia, focusing on the distinct sound patterns of Australian PTLs.

The vision-based detection employs YOLO algorithms \cite{yolo} for traffic light detection, followed by pixel counting for feature extraction. The audio inference part extracts MFCC \cite{mfccreview} to be used as features, a widely recognized method in audio classification. The signal length of our audio segments were determined after extensive comparisons of various lengths. Finally, the system combines the visual and auditory features within the audio signal length, employing feature level fusion. A machine learning model trained on combined audio-visual features is used for final classification. This model is aimed to significantly enhance the robustness and accuracy of traffic light classification in dynamic urban environments.

\begin{figure}[ht]
\centering
\includegraphics[width=1\linewidth]{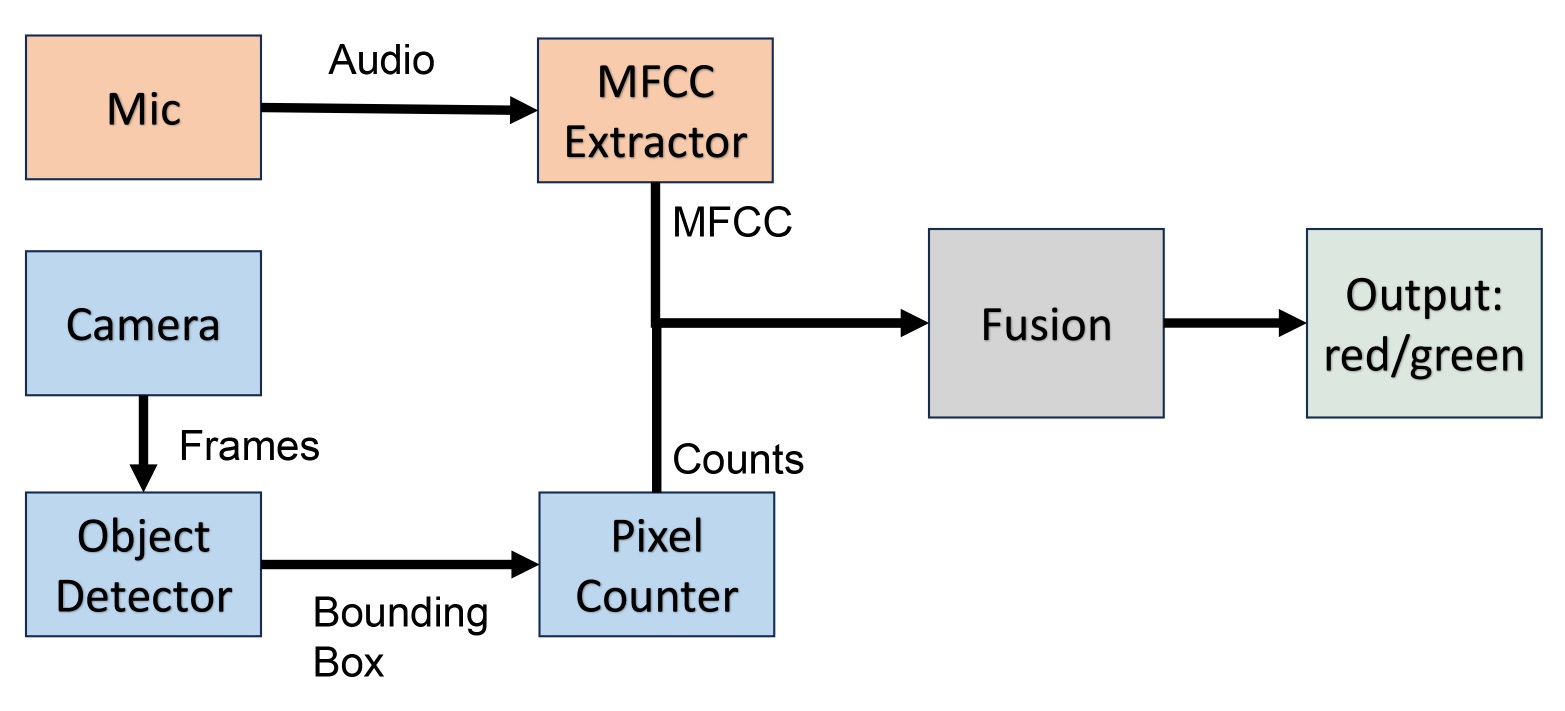}
\caption{Proposed Audio-Visual Feature Fusion Architecture}
\label{fig:sysarch}
\end{figure}

As shown in Figure \ref{fig:sysarch}, the system processes vision and audio data synchronously. It extracts information from each segment of the audio frame, combining these inputs for the final classification decision.

\begin{figure}[ht]
    \centering
    \begin{subfigure}[b]{0.232\textwidth}
        \centering
        \includegraphics[width=\textwidth]{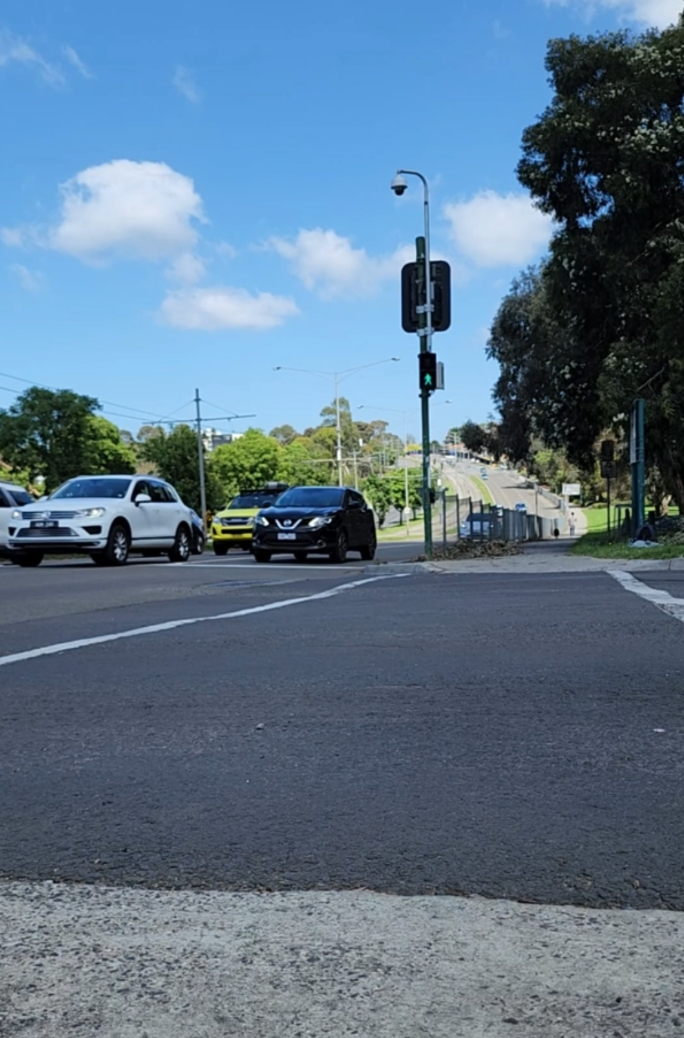}
        \caption{Containing 'green' datapoint}
        \label{fig:videoimagea}
    \end{subfigure}
    \begin{subfigure}[b]{0.24\textwidth}
        \centering
        \includegraphics[width=\textwidth]{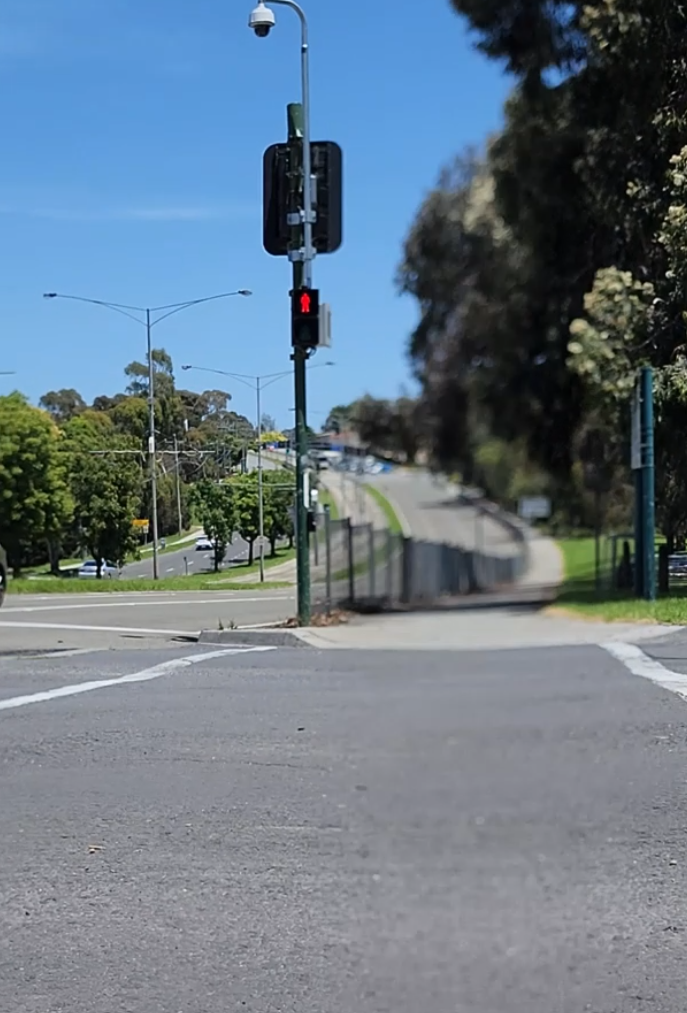}
        \caption{Containing 'red' datapoint}
        \label{fig:videoimageb}
    \end{subfigure}
    \caption{Examples from our video dataset collected onboard Unitree Go1 robot using smartphone camera}
    \label{fig:videobothimages}
\end{figure}

Testing of our approaches involved a dataset captured by a Unitree Go1 quadruped robot equipped with a smartphone. This dataset encompassed scenarios with different degrees of visual occlusion and robot motion, including 50 minutes of unobstructed footage on a stationary robot, 15 minutes of full or partial visual occlusion, and 20 minutes with the robot in motion, as depicted in figure \ref{fig:videobothimages}. The testing data also included the robot producing motor and footstep sounds while in motion, presenting an additional challenge for audio classification due to potential interference with PTL audio patterns. These tests were crucial in assessing the model's robustness and adaptability under real-world urban conditions.

The system's performance is evaluated by comparing the outputs of the best-performing vision, audio, and fusion detection algorithms. Each data point used for this comparison corresponds to the length of the audio frame. Key performance indicators include accuracy for green and red classifications against timestamps, and average inference time per data point. Our video dataset operates at 30 FPS with a resolution of 1920x1080, aligning with the majority of consumer-grade cameras.

\section{Vision-based Detection}
\label{sec:vision}

PTLs in images of crosswalks typically represent only a minor fraction of the total pixels. Traffic light state detection has been extensively implemented in literature using various methods. Authors in \cite{8569683} have proposed single shot detection that uses the same model for PTL detection and state classification. Authors in \cite{Wang2018DesignOT, 8434400, 9990723} utilize a combination of object detection and color classification. 

The ImVisible dataset \cite{yu2019lytnet} provides images of intersections in Shanghai, China along with annotations for crosswalks, in the form of coordinates that mark the start and end points of the midline of zebra crossings on the original image of size 4032x3024. These annotations also include binary information about whether the crosswalk is obstructed.  This particular dataset encompasses a range of crosswalks in Shanghai under various conditions of weather, daylight, and traffic.  There is extensive similarity between Shanghai and Australian PTLs in terms of shape, structure, and color. We chose to exclude images from intersections featuring countdown timers within the PTL housing, resulting in a refined dataset comprising 1477 images with red PTLs, 1303 with green PTLs, and 412 without any PTLs. This is done to align with Australian PTLs which do not feature a countdown timer. To facilitate PTL detection training, we found it necessary to manually annotate the PTL housings. For this purpose, we employed YOLO-accelerated annotation via RoboFlow, complemented with manual verification to ensure accuracy. The completed annotations are in YOLO format, including the binary class ID (red/green) and normalized coordinates.

Building upon the work in \cite{Ngoc2023OptimizingYP}, which evaluated the performance of various YOLO models (YOLOv5 to YOLOv8) in traffic light signal detection—where YOLOv8 emerged as the top performer —we aimed to compare the fine-tuning of YOLO's latest model, YOLOv8n, for detecting red and green PTLs against two approaches: training YOLO on multiple classes (red, green) and training YOLO on single class (PTL). YOLO fine-tuned on red/green PTL classes can allow the resulting model to perform single shot classification. However, YOLO fine-tuned on single class (PTL) will need further classification to determine final light state. It's important to note that we used the same dataset for both single and multi-class training, which inherently means less data per class in multi-class scenarios. 

\begin{figure}
    \centering
    \includegraphics[width=1\linewidth]{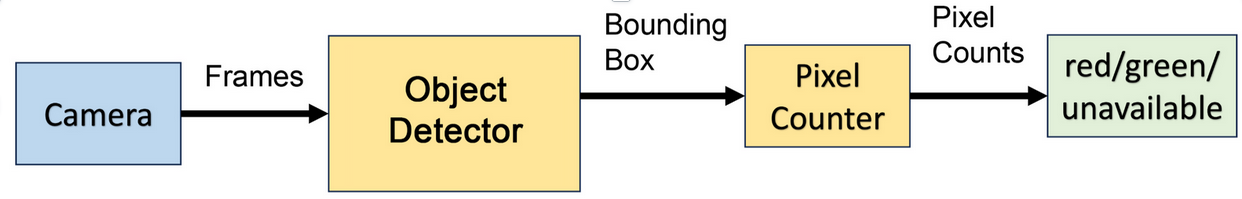}
    \caption{PTL state detection using object detection and pixel counting}
    \label{fig:videoarch}
\end{figure}

The single-class object detection method we adopted involves analyzing the red or green pixel counts within the bounding box. For this, we use pixel counting approach based on the Hue-Saturation-Value (HSV) color scale. Since we are using object detection for our vision methods, there is a possibility that no object is detected within an image. In that case, our method is unable to make a final classification, thus, the output of our classification will be 'Unavailable'. Hence, only three outputs are possible from all our methods: red, green, or unavailable. Our pixel counting method is illustrated in figure \ref{fig:videoarch}.

The Red-Green-Blue (RGB) color model is a prevalent method for image representation, which allows for segmentation of light value through distinct red, green, and blue components. By creating an RGB mask, one can compare the R, G, and B channels. However, the RGB model is sensitive to variations in lighting, leading to potential difficulties in detecting varying shades of green or red \cite{9109411}. To address this limitation, an alternative color space, known as HSV, is employed alongside RGB to classify traffic lights by hue value in \cite{9109411, Wang2018DesignOT, 8434400}. 

HSV is particularly effective in distinguishing color values under diverse lighting conditions due to its separation of the color component (hue) from the color intensity (saturation) and brightness (value). This method mirrors the approach used in \cite{Wang2018DesignOT} for classifying vehicular traffic lights (VTLs) in Hangzhou, China, where the hue ranges for red and green VTLs were identified as 160-179 and 40-70, respectively. Our approach intends to chart hue values from our dataset and establish unique hue ranges for each PTL class. Using only hue (H) values allows for distinction of color value in various levels of saturation and lighting, which are represented by the saturation (S) and value (V) values respectively. Devices for capturing images, including the smartphone used for our data collection, typically utilize the RGB color model.

%





\begin{figure}[ht]
    \centering
    \includegraphics[width=0.9\linewidth]{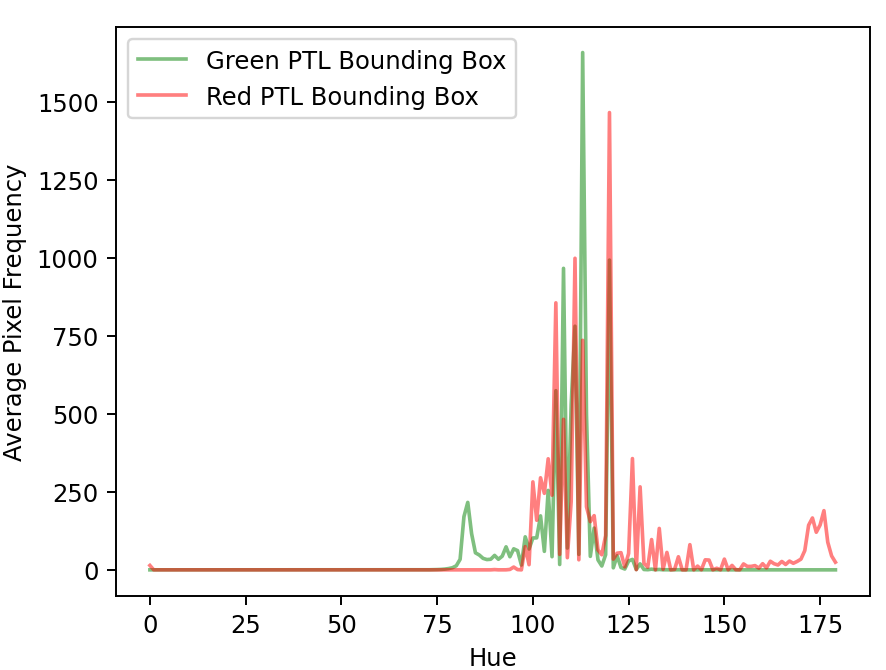}
    \caption{Hue histogram of red and green PTL bounding boxes, averaged over all frames in our video dataset containing no occlusion captured on-board robot.}
    \label{fig:hue_a}
\end{figure}

Upon completing the annotation of the ImVisible dataset \cite{yu2019lytnet} and fine-tuning YOLOv8n for single-class operation, our methodology involves classifying PTLs as red or green by comparing the percentages of red and green pixels within their bounding boxes against the predetermined hue ranges. To determine these optimum hue ranges, we utilized our single-class YOLO model to detect bounding boxes in our robot's dataset, which comprises 50 minutes of footage without occlusions, captured using a stationary robot setup. In addressing the class imbalance in our dataset—1869 green data points versus 1020 red ones—we limited our hue range calculation to the first 1020 green data points. These bounding boxes were then converted to the HSV color space, and the pixel count for each hue value was averaged and graphically represented in figure \ref{fig:hue_a}.

Our color analysis of the video dataset yielded distinct hue ranges for green and red PTL bounding boxes, identified as 75-100 for green and 170-180 for red. These findings suggest that these hue ranges are representative of the unique color characteristics inherent to each PTL class. Interestingly, these identified ranges bear similarity to those used in previous research \cite{Wang2018DesignOT}, although they deviate slightly from the hues associated with VTLs. This variation in hue values between PTLs and VTLs could potentially be attributed to the differences in coloration as well as regional variances in traffic light design and manufacturing.

\begin{figure}[ht]
    \centering
    \begin{subfigure}[b]{0.11\textwidth}
        \centering
        \includegraphics[width=\textwidth]{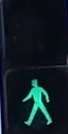}
        \caption{Green PTL Bounding Box}
        \label{fig:ptlgreeninner}
    \end{subfigure}
    \hfill 
    \begin{subfigure}[b]{0.11\textwidth}
        \centering
        \includegraphics[width=\textwidth]{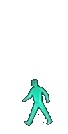}
        \caption{Pixels within hue range}
        \label{fig:ptlgreenpixels}
    \end{subfigure}
    \hfill
    \centering
    \begin{subfigure}[b]{0.11\textwidth}
        \centering
        \includegraphics[width=\textwidth]{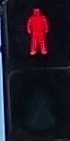}
        \caption{Red PTL Bounding Box}
        \label{fig:ptlredinner}
    \end{subfigure}
    \hfill 
    \begin{subfigure}[b]{0.11\textwidth}
        \centering
        \includegraphics[width=\textwidth]{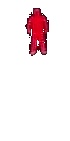}
        \caption{Pixels within hue range}
        \label{fig:ptlredpixels}
    \end{subfigure}
    \caption{Extracting pixels within hue range of PTL bounding boxes from example frames}
    \label{fig:ptlhsvgreen}
\end{figure}

The hue-based filtering results for bounding boxes for an example data point (figures \ref{fig:ptlgreeninner} and \ref{fig:ptlredinner}) are shown in figures \ref{fig:ptlgreenpixels} and \ref{fig:ptlredpixels}. Our observations from this exercise revealed that the designated hue ranges effectively encapsulate the unique colors of the PTLs. Furthermore, we deduced that if a bounding box's image contains a higher concentration of pixels within a specific hue range, it is feasible to classify the PTL accordingly, based on the predominant color range present. This approach demonstrates a reliable method for differentiating between red and green PTLs, leveraging the distinct color properties captured within the hue ranges. 

For classifying using hue thresholds, Let's define:
\begin{itemize}
    \item \( R \) as the number of red pixels within the bounding box.
    \item \( G \) as the number of green pixels within the bounding box.
\end{itemize}

Then, the percentage of red pixels (\( P_{\text{red}} \)) and green pixels (\( P_{\text{green}} \)) can be calculated as follows:

\[
P_{\text{red}} = \frac{R}{R + G} \times 100
\]
\[
P_{\text{green}} = \frac{G}{R + G} \times 100
\]

\( P_{\text{red}} \) and \( P_{\text{green}} \) indicate the proportion of pixels classified as green or red, adjusted for varying resolutions. The final classification \( C \) is determined by comparing \( P_{\text{red}} \) and \( P_{\text{green}} \):

\[
C = 
\begin{cases} 
  \text{'Unavailable'}, & \text{if } P_{\text{red}} = P_{\text{green}} \text{ or no detections} \\
  \text{'Red'}, & \text{if } P_{\text{red}} > P_{\text{green}} \\
  \text{'Green'}, & \text{if } P_{\text{red}} < P_{\text{green}}
\end{cases}
\]
We conducted a comparative analysis of two approaches: object detection for PTL followed by hue-based classifier, and an object detection model fine-tuned for red and green PTL detection. The aim was to assess the final classification accuracy of these approaches. We evaluated the performance of our vision inference methods using all images from ImVisible dataset \cite{yu2019lytnet} as well as all frames from our video dataset captured onboard the robot. We only use data points without visual occlusion and without robot movement to analyze accuracy of our strategies in ideal conditions. This exercise contained 1303 images containing green PTLs, 1477 images containing red PTLs from the ImVisible dataset, which was used for training. From our video dataset, we extracted 56,070 frames containing green PTL and 30,780 frames containing red PTL for a comprehensive evaluation of unseen data. Since all of the data points contain either a red or green PTL, an 'unavailable' classification is considered inaccurate in case no detections are made. We firstly split images from our two datasets into red or green sets, and analyze accuracy of our models against all elements in the set, tabulated in table \ref{tab:vision_methods}. 

\begin{table}[h!]
\centering 
\caption{Classification accuracy of vision-based approaches on ImVisible dataset \cite{yu2019lytnet} and all frames without visual occlusion from video dataset captured on-board robot}
\begin{tabular}{|l|c|c|c|}

\hline
\begin{tabular}[c]{@{}c@{}}Classification Method \end{tabular} & \begin{tabular}[c]{@{}c@{}}ImVisible\\Dataset \cite{yu2019lytnet}\end{tabular} & 
\begin{tabular}[c]{@{}c@{}}Robot Video\\Dataset\end{tabular} \\ 
\hline
\begin{tabular}[c]{@{}c@{}}Single class YOLO + HSV thresholding\\ \end{tabular} & \textbf{98.4\%} & \textbf{96.7\%} \\
\hline
\begin{tabular}[c]{@{}c@{}}Multi class YOLO\end{tabular} & \text{97.1\%} & \text{90.5\%} \\
\hline
\end{tabular}
\label{tab:vision_methods}
\end{table}

From our analysis, it is revealed that using HSV thresholds proved to be most effective on both datasets, achieving 96.7\% accuracy on our video dataset. Comparatively, YOLO fine-tuned on red/green PTL classes performed worse on the dataset it was trained on, but was able to achieve 90.5\% accuracy on our video dataset. For integration in our fusion model, we elect our HSV thresholding approach coupled with YOLO trained on a single class due it its out-performance of single shot detection through YOLO.

\section{Audio-based Detection}
\label{sec:audio}

The integration of vision and audio data in our system necessitates careful consideration of synchronization, audio frame length, and the selection of features that accurately reflect the state of PTL. Our model's performance was evaluated using audio clips of varying lengths: 250ms, 500ms, 750ms, and 1000ms. We found that the audio signals emitted from PTLs typically repeat every 100-200ms, about six times per second. Notably, a delay occurs when the light changes to green, marked by an initial sound effect before the standard pattern begins, resulting in a variable audio delay of 300-600ms during the transition from red to green.

\subsection{Number of Mel Frequency Cepstral Coefficients}

MFCC are a cornerstone in sound event and speech recognition, representing the short-term power spectrum of sound \cite{mfccreview}. These coefficients are derived from a linear cosine transform of a log power spectrum on a nonlinear mel scale. The choice of the number of MFCC and audio frame length significantly influences classification accuracy \cite{dave2013feature}. While traditionally, 13 cepstral coefficients are used, higher order coefficients are often excluded as they represent rapid changes in estimated energies and contain less information \cite{gupta2013feature}. The 0th order coefficient, indicating audio volume, is typically disregarded. Derivatives of MFCC, known as delta and delta-delta coefficients, are sometimes included to represent signal variation \cite{GHIURCAU2012829}. Research by \cite{8073600} found that 16 MFCC classified using a k-Nearest Neighbor (k-NN) classifier was optimal for an intruder detection system. To get optimum number of MFCC, we mirror the approach in \cite{8073600}, which compared widely used classifiers against a range of MFCC. For our study, we opt for a cross-comparison of two classifiers:

\begin{itemize}
\item k-Next Neighbors (k-NN): Implements various distance measures and efficient neighbor search in large datasets \cite{guo2003knn}.
\item Random Forest (RF): Uses random vectors to grow an ensemble of trees for voting on the most popular class \cite{biau2016random}.
\end{itemize}

We also compared these classifiers against Bayesian Networks \cite{koski2011bayesian} but the results were significantly worse, hence we elected to present results from our best performing classifiers. Given the unique strengths of each classifier and considering variables like environmental noise and audio pattern variability, we propose a cross-comparison across N MFCC, with N ranging from 10, 12, 14, 16, 18, 20, 24 and 28. This approach aligns with studies like \cite{8073600, chen2009speaker}. We also examine the impact of including delta and delta-delta coefficients, doubling the feature count for delta and tripling for delta and delta-delta.

\begin{figure}[ht]
    \centering
    \includegraphics[width=1\linewidth]{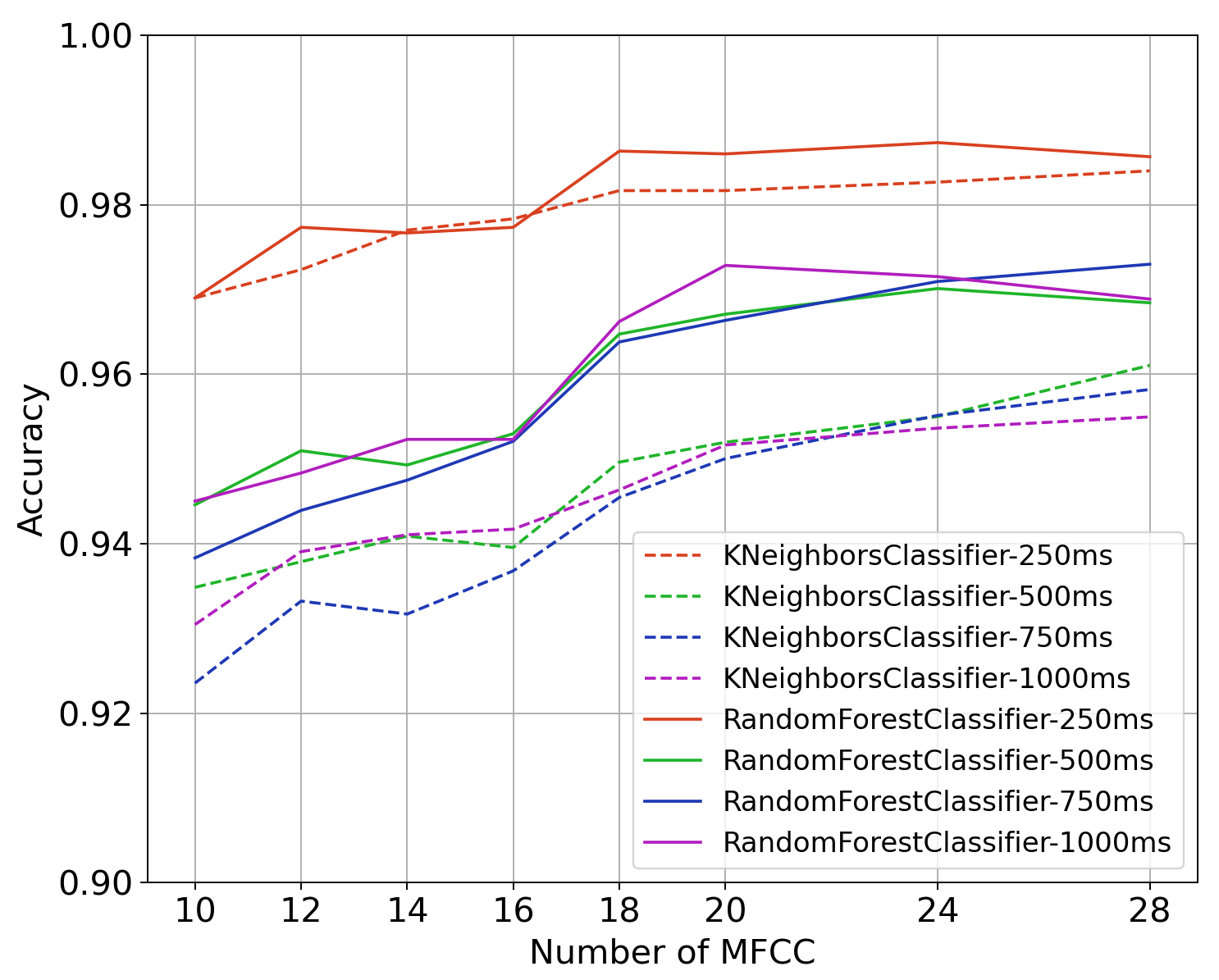}
    \caption{Cross-Comparison of Random Forest \cite{biau2016random} and k-Nearest Neighbor \cite{guo2003knn} classifiers trained on varying audio lengths and number of MFCC for red/green classification.}
    \label{fig:mfcc_cross}
\end{figure}

Our self-collected audio dataset comprises 2 hours of green PTL data and 2 hours of red PTL data recorded using a handheld smartphone. We extracted 30\% of our audio data to be used for testing and used the remainder for training. The audio signals were split according to frame length, and features were extracted from all N MFCC. These features were then trained on the two classifiers, and their accuracies were charted in figure \ref{fig:mfcc_cross}.

The Random Forest classifier trained on 250ms audio frames with 24 MFCC achieved the highest accuracy at 98.7\%. 

\subsection{Including delta and delta-delta coefficients}

Repeating the experiment with delta and delta-delta coefficients added to each N, the highest accuracy (98.6\%) was achieved using 24 MFCC (plus 24 derivatives) in 250ms frames with the Random Forest Classifier. The inclusion of delta-delta coefficients (24 + 48 features) also yielded 98.6\% accuracy using the same model and parameters. Despite the close performance, we decided against including derivatives in our coefficients as not using derivates still performed marginally better.

%


\subsection{Choice of Frame Length and Algorithm}

With 24 MFCC identified as optimal, we extracted accuracy data for our classifiers trained on this feature set across varying audio frame lengths. The results, presented in Table \ref{audioframelength}, indicate the highest accuracy with 250ms frames. Notably, the Random Forest classifier outperformed k-NN across all frame lengths, leading us to select the Random Forest classifier trained on 250ms frames with 24 MFCC as our model of choice. 

\begin{table}[ht]
\centering
\begin{tabular}{|l|c|c|}
\hline
\text{Classifier Type}        & \text{Frame Length} & \text{Accuracy} \\  \hline
 & 250ms          & 98.3\%    \\ \cline{2-3} 
KNeighbors                     & 500ms          & 95.5\%    \\ \cline{2-3} 
    Classifier                 & 750ms          & 95.5\%    \\ \cline{2-3} 
                     & 1000ms         & 95.4\%    \\ \hline
 & \textbf{250ms}         & \textbf{98.7\%}    \\ \cline{2-3} 
\text{Random Forest}                       & 500ms         & 97.0\%    \\ \cline{2-3} 
            \text{Classifier}           & 750ms         & 97.1\%    \\ \cline{2-3} 
                       & 1000ms        & 97.2\%    \\ \hline
\end{tabular}
\caption{Comparison of Random Forest \cite{biau2016random} and k-Nearest Neighbor \cite{guo2003knn} classifiers trained on varying audio lengths and 24 MFCC for red/green classification.}
\label{audioframelength}
\end{table}


\subsection{Performance on Video Dataset}

Testing our audio classifier against the robot-captured test dataset, we compiled the accuracies in Tables \ref{tab:cross_a}, \ref{tab:cross_b}, and \ref{tab:cross_c}. The average processing time for 250ms of audio was 4ms. Our audio classification method demonstrated high accuracy (97.0\%) in scenarios with no occlusion and a stationary vision feed. In cases with visual occlusion, the accuracy stood at 97.8\%. While the robot was in motion, the audio model distinguished between the robot's movements and the green light's audio pattern, accurately detecting red lights. However, the robot's movements affecting the green light's audio pattern led to a lower accuracy for green light detection. The audio model maintained an accuracy of 90.4\% while classifying during robot locomotion.

\section{Audio-Visual Fusion}
\label{sec:fusion}

The human brain effortlessly processes and integrates inputs from multiple sensory modalities, such as vision and audio. In contrast, most computational models for machine perception are tailored to single modalities, optimized specifically for unimodal applications. However, the fusion of data from disparate sources can significantly enhance accuracy, as evidenced in studies like \cite{Zhou2023MTANetMN}. This particular research successfully fused thermal (RGB-T) and RGB images to improve urban scene understanding, a critical advancement for assisted driving and autonomous vehicles. Generally, multimodal fusion in machine perception can be categorized into two primary types: decision-level and feature-level fusion \cite{atrey2010multimodal, ortega2019multimodal}. One notable application of audio-video fusion is in the field of deepfake detection. The study in \cite{muppalla2023integrating} combined features from a face detection model with MFCC, relying on just one frame per second for visual inference due to the low likelihood of significant change in the subject's appearance within such a short interval. Another approach, presented in \cite{Ozay2009EnsembleDA}, proposed a decision-level fusion framework within a multilayer hierarchical ensemble learning architecture. This structure allows for generative discriminative modeling, efficient probability density estimations, and reduction of data entropy across vector spaces, making it highly suitable for object classification and target detection. Furthermore, \cite{Jia2022AME} demonstrated the use of audio and video features alongside motion capture for emotion recognition, employing decision-level fusion. \cite{NEURIPS2021_76ba9f56} introduced a transformer-based architecture that utilizes 'attention bottlenecks' for effective multimodal fusion at various layers. This method has substantially improved machine perception while simultaneously reducing computational demands. In our research, we aim to explore and compare the effectiveness of these two fusion strategies—feature-level and decision-level fusion—for our classifier. This comparison will provide insights into the most efficient and accurate methods for integrating multimodal data in machine perception, particularly in contexts where both audio and visual cues are crucial for accurate interpretation. 

In this study, we perform a comparison of fusing audio and vision features for classification versus combining the decisions of audio and vision-based classifiers to output classification at the decision level. From the insights gained through our audio method, we have established 250ms as the optimal frame length for audio processing. In our fusion approach, we define a single data point as encompassing 250ms of both video frames and audio. The fusion logic we adopt will be based on the most effective method identified from our earlier comparisons of vision and audio inference. The approach in \cite{muppalla2023integrating} suggests using only one frame per data point, as the presence of a PTL is unlikely to change rapidly. However, our system's efficient inference time allows for the analysis of multiple observations within this time frame.

\begin{figure}[h!]
    \centering
    \includegraphics[width=1\linewidth]{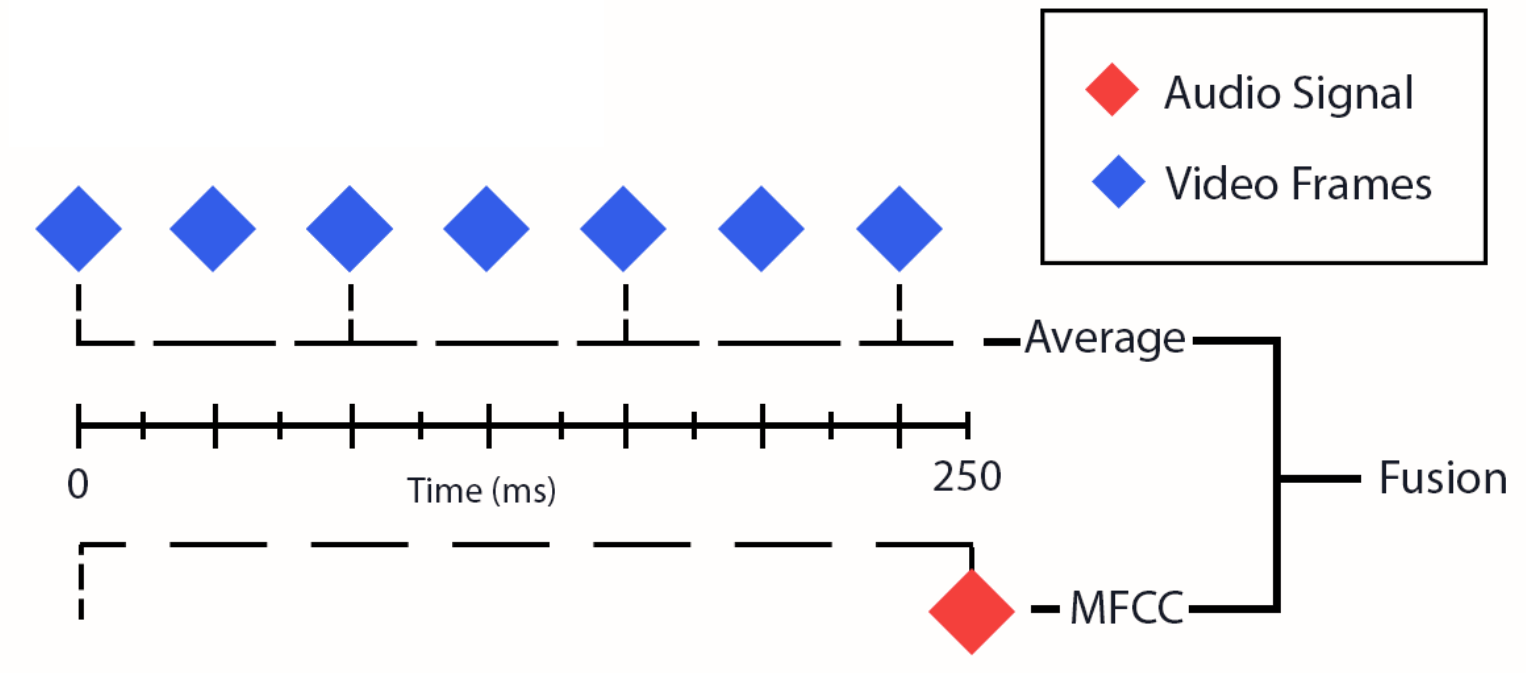}
    \caption{Audio-Visual feature extraction and synchronization before fusion}
    \label{fig:time-series}
\end{figure}

Given that the average inference time for our YOLO model is 62ms and a video frame appears every 33.3ms when the FPS is 30, we aim to synchronize our fusion methods by pairing 250ms of audio frames with corresponding vision frames. Our video, captured at 30 FPS is firstly divided into segments corresponding to 250ms audio frames, resulting in 7.5 frames per segment. Consequently, we utilize the first 7 frames for each 250ms audio segment. Analyzing all 7 frames within a 250ms segment would take approximately 434ms. Hence, we can only use a maximum of 4 frames, equalling an estimated 248ms of inference time. Since our capability for inference is approximately half of the video frames, we average the features from every other frames within the audio clip's duration to represent the video segment's features.  By selecting alternating frames, we effectively focus on 4 frames for each data point. For each segment, the audio features comprise 24 MFCC extracted from the corresponding audio segment. The overall process of this system for each data point is illustrated in figure \ref{fig:time-series}.

\subsection{Feature Level Fusion}

In our study, we developed an integrated approach for feature-level fusion by seamlessly combining audio and visual features that represent the state of PTL. This process involves concatenating the audio features with visual attributes and then reshape these features to prepare it for both training and inference processes. For training, we concatenate MFCC features from our handheld audio dataset with pixel percentage features from the ImVisible dataset \cite{yu2019lytnet}, incorporating 1303 images of green pedestrian traffic lights (PTLs), 1477 images of red PTLs, and 412 images without PTLs. This inclusion of non-PTL images is deliberate, aimed at training the model to depend on audio inputs in the absence of visual detections. A Random Forest classifier \cite{biau2016random} is subsequently trained on these combined features.

For extracting vision features, we use [\( P_{\text{red}} \), \( P_{\text{green}} \)] that are color percentage values calculated from pixel counts for red and green hue values. These features ensure that the pixel ratio remains consistent for both training and evaluation, independent of the variations in input image resolution. Given that our vision-based classifier is capable of identifying three distinct outcomes (red, green, or unavailable), it provides a broader spectrum of information in comparison to the audio classifier, which discerns only between red and green with corresponding confidences. To accurately represent these three potential visual outcomes as distinct features, we default to using [0,0] for scenarios where no detection occurs. Thus, based on the proportion of high (\( x \)) and low (\( y \)) pixel ratios, our features are delineated as either [\( x \),\( y \)], [\( y \),\( x \)], or [0,0]. For classification, we average these features across all frames selected for analysis. The visual attributes are then combined with the audio features, which consist of 24 MFCC extracted from 250 milliseconds of audio. Random Forest \cite{biau2016random} is then used to perform binary classification of these features into red/green.

\subsection{Decision Level Fusion}

For decision level fusion, we sum the confidence scores from our vision and audio classification methods for each data class, with the predicted class being the one with the highest cumulative confidence. In terms of vision, the confidence metric is derived from the average detection confidence of the bounding boxes over the 4 frames. For audio, it is based on the class confidence for the respective data class. This method relies on combining the individual strengths of audio and vision classifiers by considering their respective confidence levels in predicting the PTL state, thereby leveraging the advantages of both modalities to enhance overall classification accuracy. When our object detector is unable to make a detection, only audio confidences are used as we don't have confidences for bounding boxes.

\section{Results}
\label{sec:results}

The classification accuracy of both our feature level fusion and decision level fusion methods against red and green data points has been thoroughly analyzed and documented in Tables \ref{tab:cross_a}, \ref{tab:cross_b}, and \ref{tab:cross_c}. The methods were analyzed against our video dataset captured onboard our robot. These tables detail the overall accuracy of each fusion strategy under three distinct conditions: when the robot is stationary when it is in motion, and when the camera view is occluded. Our fusion model, handling 250ms of vision and audio data, averaged an inference time of 242ms for processing 4 frames at a resolution of 1920x1080, using the Nvidia RTX 4060 GPU.

\subsection{Under no visual occlusion on stationary robot}
Analysis for datapoints containing no visual occlusion and no robot noise is presented in table \ref{tab:cross_a}. We observed a notable improvement in accuracy compared to the single modality approaches. Specifically, the feature level fusion method achieved an accuracy of 99.1\%, while the decision level fusion method reached 98.3\%. It is interesting to note that while the decision-level fusion’s accuracy was slightly below the vision-only approach (98.5\% accurate), the integration of features for training and inference in feature-level fusion exhibited superior performance.


\begin{table}[h!]
\centering 
\caption{Classification Accuracy of Audio, Vision and Fusion Approaches. Data: Captured on-board robot. Type: No visual occlusion. Green Light Data Points: 7476, Red Light Data Points: 4104, Overall Data Points: 11580}
\begin{tabular}{|l|c|c|c|}
\hline
\begin{tabular}[c]{@{}c@{}}Classification Method\\ \end{tabular} & \begin{tabular}[c]{@{}c@{}}Green\\ Light\\ Accuracy\end{tabular} & \begin{tabular}[c]{@{}c@{}}Red\\ Light\\ Accuracy\end{tabular} & \begin{tabular}[c]{@{}c@{}}Overall\\ Accuracy\end{tabular} \\ 
\hline
\begin{tabular}[c]{@{}c@{}}Vision-only\\ \end{tabular} & \text{99.3\%} & \text{97.1\%} & \text{98.5\%} \\
\hline
\begin{tabular}[c]{@{}c@{}}Audio-only\\\end{tabular} & \text{98.1\%} & \text{95.0\%} & \text{97.0\%} \\
\hline
\begin{tabular}[c]{@{}c@{}}Vision + Audio, Feature-level\end{tabular} & \textbf{99.7\%} & \textbf{98.0\%} & \textbf{99.1\%} \\
\hline
\begin{tabular}[c]{@{}c@{}}Vision + Audio, Decision-level\\  \end{tabular} & \text{99.2\%} & \text{96.6\%} & \text{98.3\%} \\ 
\hline
\end{tabular}
\label{tab:cross_a}
\end{table}


\subsection{Under Visual Occlusion}
Analysis for datapoints containing  visual occlusion is presented in table \ref{tab:cross_b}. In scenarios of complete or partial visual occlusion, vision only method performs very poorly as it is unable to make detections, and made accurate predictions for only 4.9\% of the datapoints. The feature level fusion method demonstrated an accuracy of 95.6\%, compared to 97.4\% for the decision level fusion. This finding indicates that decision-level fusion is more advantageous in situations where visual data is not available, as it relies more on audio confidence values for classification. Our audio classification model achieved the highest accuracy, 97.8\%, slightly higher than our decision-level fusion.


\begin{table}[h!]
\centering 
\caption{Classification Accuracy of Audio, Vision and Fusion Approaches. Data: Captured on-board robot. Type: Visual occlusion. Green Light Data Points: 3256, Red Light Data Points: 1812, Overall Data Points: 5068}
\begin{tabular}{|l|c|c|c|}
\hline
\begin{tabular}[c]{@{}c@{}}Classification Method\\ \end{tabular} & \begin{tabular}[c]{@{}c@{}}Green\\ Light\\ Accuracy\end{tabular} & \begin{tabular}[c]{@{}c@{}}Red\\ Light\\ Accuracy\end{tabular} & \begin{tabular}[c]{@{}c@{}}Overall\\ Accuracy\end{tabular} \\ 
\hline
\begin{tabular}[c]{@{}c@{}}Vision-only\\ \end{tabular} & \text{5.4\%} & \text{4.0\%} & \text{4.9\%} \\
\hline
\begin{tabular}[c]{@{}c@{}}Audio-only\\\end{tabular} & \textbf{99.7\%} & \textbf{94.3\%} & \textbf{97.8\%} \\
\hline
\begin{tabular}[c]{@{}c@{}}Vision + Audio, Feature-level\end{tabular} & \text{98.1\%} & \text{91.2\%} & \text{95.6\%} \\
\hline
\begin{tabular}[c]{@{}c@{}}Vision + Audio, Decision-level\\  \end{tabular} & \textbf{99.7\%} & \text{93.4\%} & \text{97.4\%} \\ 
\hline
\end{tabular}
\label{tab:cross_b}
\end{table}

\subsection{Under Robot Movement}
Analysis for datapoints containing  the robot in motion is presented in Table \ref{tab:cross_c}. Feature-level fusion showed enhanced performance in conditions where the robot was moving, affecting both visual and audio inputs. Trained on a combination of vision and audio data, feature level fusion method is better equipped to manage the most challenging category. As a result, for data points where the robot is moving, feature level fusion achieves a higher accuracy of 98.8\%, whereas decision level fusion shows a slightly lower accuracy, correctly classifying 96.0\% of the data points. These results highlight the effectiveness of fusion strategies in diverse and challenging operational environments, underlining their importance in enhancing the robustness and accuracy of multimodal perception systems.

\begin{table}[h!]
\centering 
\caption{Classification Accuracy of Audio, Vision and Fusion Approaches. Data: Captured on-board robot. Type: Robot moving. Green Light Data Points: 3136, Red Light Data Points: 1716, Overall Data Points: 4852}
\begin{tabular}{|l|c|c|c|}
\hline
\begin{tabular}[c]{@{}c@{}}Classification Method\\ \end{tabular} & \begin{tabular}[c]{@{}c@{}}Green\\ Light\\ Accuracy\end{tabular} & \begin{tabular}[c]{@{}c@{}}Red\\ Light\\ Accuracy\end{tabular} & \begin{tabular}[c]{@{}c@{}}Overall\\ Accuracy\end{tabular} \\ 
\hline
\begin{tabular}[c]{@{}c@{}}Vision-only\\ \end{tabular} & \text{96.9\%} & \text{95.7\%} & \text{96.5\%} \\
\hline
\begin{tabular}[c]{@{}c@{}}Audio-only\\\end{tabular} & \text{89.7\%} & \text{91.7\%} & \text{90.4\%} \\
\hline
\begin{tabular}[c]{@{}c@{}}Vision + Audio, Feature-level\end{tabular} & \textbf{99.6\%} & \text{97.3\%} & \textbf{98.8\%} \\
\hline
\begin{tabular}[c]{@{}c@{}}Vision + Audio, Decision-level\\  \end{tabular} & \text{94.6\%} & \textbf{98.5\%} & \text{96.0\%} \\ 
\hline
\end{tabular}
\label{tab:cross_c}
\end{table}


\section{Robot Implementation}
\label{sec:demo}
The proposed system for detecting the state of PTLs is implemented on a Unitree Go1 robot. This robot is fitted with a smartphone mounted on its front, which captures video footage. The system's operation is supported by a laptop equipped with an RTX 4060 GPU. The robot's Wi-Fi network is connected with the video-capture smartphone as well as the laptop. This setup enables access to the robot's Robot Operating System (ROS) network, a crucial component for the system's functionality.

Central to the system's operation is a FastAPI web server running on the laptop, which, in conjunction with a Python video processing module, leverages the capabilities of our proposed fusion model. The web server is programmed to request rear-facing video footage from the client device at a resolution of 1920x1080, capturing at a frame rate of 30 FPS every 250 milliseconds. This video data is then analyzed by the video processing module, which performs feature extraction for pixel counting. It averages these features and integrates them with 24 MFCC extracted from the video's audio track. 

The decision for the features, inferred using our fusion model, alongside the dimensions of the detected bounding box, is published to a ROS topic named /traffic\_light\_state. A ROS node, subscribed to this topic, interprets the information to determine the robot's next actions. If a green light is detected, the system uses the bounding box's size to estimate the distance to a designated goal point, leveraging the robot's odometry data for accuracy. Upon determining the goal point's location, the ROS node guides the robot toward it, ensuring it maintains the correct orientation and path. As the robot nears the goal point, within a distance of one meter, it receives a command to stop, indicating successful navigation across the pedestrian crossing.

\section{Conclusion}

In this study, we introduce an audio-visual fusion model for detecting pedestrian traffic light (PTL) states from the view of a quadruped robot. Our vision model utilizes established vision-based detectors for initial PTL identification and incorporates a simple pixel-counting approach for determining the state as red or green. The audio component of our model extracts Mel-Frequency Cepstral Coefficients (MFCC) features \ref{fig:mfcc_cross}, integrating these with visual data at a feature level. This fusion technique can handle data from multiple frames within a set timeframe, enhancing the model's adaptability and performance.

Our fusion method notably achieved a classification accuracy exceeding 95\% when the robot's view is under visual partial or full occlusions, substantially surpassing vision-only solutions, albeit slightly behind the audio-only modality. In dynamic scenarios where the robot is in motion, our fusion model performed the best over vision-only and audio-only modalities with over 98\% accuracy. With an average inference time of 64ms on a standard consumer-grade GPU, our model is well-suited for real-time processing, meeting a critical requirement in urban robotics.

There is room for improvement for this work. Both the vision and audio components of our model offer room for optimization and merit comparison against alternative methods. The vision-based pixel counting, in particular, holds the potential for greater resistance to PTL design variations, like different pedestrian symbols—a hypothesis reserved for subsequent investigation. Moreover, the dataset, sourced from China and Australia, may not fully represent the vast diversity of PTL designs and environmental conditions globally. Improved hardware such as stabilized cameras and directional microphones, is also identified as an avenue for future analysis.

\balance
\bibliographystyle{plain}
\bibliography{ref.bib}

\end{document}